\documentclass[11pt,letterpaper]{article}
\usepackage{naaclhlt2016}
\usepackage{times}
\usepackage{latexsym}
\usepackage[T1]{fontenc}
\usepackage{array}
\usepackage{multirow}
\usepackage{amstext}
\usepackage{graphicx}
\usepackage{float}
\usepackage{verbatim}
\usepackage{tabularx}
\usepackage{amsmath,mathtools}
\naaclfinalcopy
\makeatletter
\providecommand{\tabularnewline}{\\}

\makeatother
\makeatletter

\newcommand{\specificthanks}[1]{\@fnsymbol{#1}}

\title{Detecting "Smart" Spammers On Social Network:\\A Topic Model Approach}
\author{Linqing Liu,\textsuperscript{1} Yao Lu,\textsuperscript{1} Ye Luo\textsuperscript{1}\textsuperscript{,\specificthanks{1}}, Renxian Zhang\textsuperscript{2}\textsuperscript{,\specificthanks{1}}, Laurent Itti\textsuperscript{1,3} and Jianwei Lu\textsuperscript{1,4}\textsuperscript{,\Thanks{Corresponding Author}}\\
	\textsuperscript{1} iLab Tongji, School of Software Enginnering, Tongji University\\
	\textsuperscript{2} Dept. of Computer Science and Technology, Tongji University\\
	\textsuperscript{3} Dept. of Computer Science and Neuroscience Program, University of Southern California\\
	\textsuperscript{4} Institute of Translational Medicine, Tongji University\\
	{\tt likicode@gmail.com}, {\tt \{95luyao,rxzhang\}@tongji.edu.cn},\\  
		 {\tt\{kennyluo2008,jwlu33\}@hotmail.com}, {\tt itti@usc.edu}}

\begin{document}
	\maketitle

\begin{abstract}
Spammer detection on social network is a challenging problem. The rigid anti-spam rules have resulted in emergence of "smart" spammers. They resemble legitimate users who are difficult to identify. In this paper, we present a novel spammer classification approach based on Latent Dirichlet Allocation (LDA), a topic model. Our approach extracts both the local and the global information of topic distribution patterns, which capture the essence of spamming. Tested on one benchmark dataset and one self-collected dataset, our proposed method outperforms other state-of-the-art methods in terms of averaged F1-score.

\end{abstract}

\section{Introduction}

Microblogging such as Twitter and Weibo is a popular social networking service, which allows users to post messages up to 140 characters. There are millions of active users on the platform who stay connected with friends. Unfortunately, spammers also use it as a tool to post malicious links, send unsolicited messages to legitimate users, etc. A certain amount of spammers could sway the public opinion and cause distrust of the social platform. Despite the use of rigid anti-spam rules, human-like spammers whose homepages having photos, detailed profiles etc. have emerged. Unlike previous "simple" spammers, whose tweets contain only malicious links, those "smart" spammers are more difficult to distinguish from legitimate users via content-based features alone ~\cite{riseofbots}.

There is a considerable amount of previous work on spammer detection on social platforms. Researcher from Twitter Inc. \cite{chu2010tweeting} collect bot accounts and perform analysis on the user behavior and user profile features. Lee et al.~\shortcite{lee2011seven} use the so-called social honeypot by alluring social spammers' retweet to build a benchmark dataset, which has been extensively explored in our paper. Some researchers focus on the clustering of urls in tweets and network graph of social spammers \cite{yang2012analyzing,wang2015making,wang2010don,yang2011free}, showing the power of social relationship features.As for content information modeling,~\cite{hu2013social} apply improved sparse learning methods. However, few studies have adopted topic-based features. Some researchers ~\cite{liu2014sdhm} discuss topic characteristics of spamming posts, indicating that spammers are highly likely to dwell on some certain topics such as promotion. But this may not be applicable to the current scenario of smart spammers.

In this paper, we propose an efficient feature extraction method. In this method, two new topic-based features are extracted and used to discriminate human-like spammers from legitimate users. We consider the historical tweets of each user as a document and use the Latent Dirichlet Allocation (LDA) model to compute the topic distribution for each user. Based on the calculated topic probability, two topic-based features, the Local Outlier Standard Score (LOSS) which captures the user's interests on different topics and the Global Outlier Standard Score (GOSS) which reveals the user's interests on specific topic in comparison with other users', are extracted. The two features contain both local and global information, and the combination of them can distinguish human-like spammers effectively.  

To the best of our knowledge, it is the first time that features based on topic distributions are used in spammer classification. Experimental results on one public dataset and one self-collected dataset further validate that the two sets of extracted topic-based features get excellent performance on human-like spammer classification problem compared with other state-of-the-art methods. In addition, we build a Weibo dataset, which contains both legitimate users and spammers.

To summarize, our major contributions are two-fold:
\begin{itemize}
		\setlength\itemsep{-0.2em}
\item We extract topic-based features (GOSS and LOSS) for spammer detection, which outperform state-of-the-art methods.
\item We build a dataset of Chinese microblogs for spammer detection.
\end{itemize}

In the following sections, we first propose the topic-based features extraction method in Section 2, and then introduce the two datasets in Section 3. Experimental results are discussed in Section 4, and we conclude the paper in Section 5. Future work is presented in Section 6.

\section{Methodology}

In this section, we first provide some observations we obtained after carefully exploring the social network, then the LDA model is introduced. Based on the LDA model, the ways to obtain the topic probability vector for each user and the two topic-based features are provided.

\subsection{Observation}
After exploring the homepages of a substantial number of spammers, we have two observations. 1) social
spammers can be divided into two categories. One is content polluters, and their tweets are all about certain kinds of advertisement and campaign. The other is fake accounts, and their tweets resemble legitimate users' but it seems they are simply random copies of others to avoid being detected by anti-spam rules. 2) For legitimate users, content polluters and
fake accounts, they show different patterns on topics which interest them.

\begin{figure}
	\includegraphics[width=1\columnwidth,height=1\columnwidth,keepaspectratio]{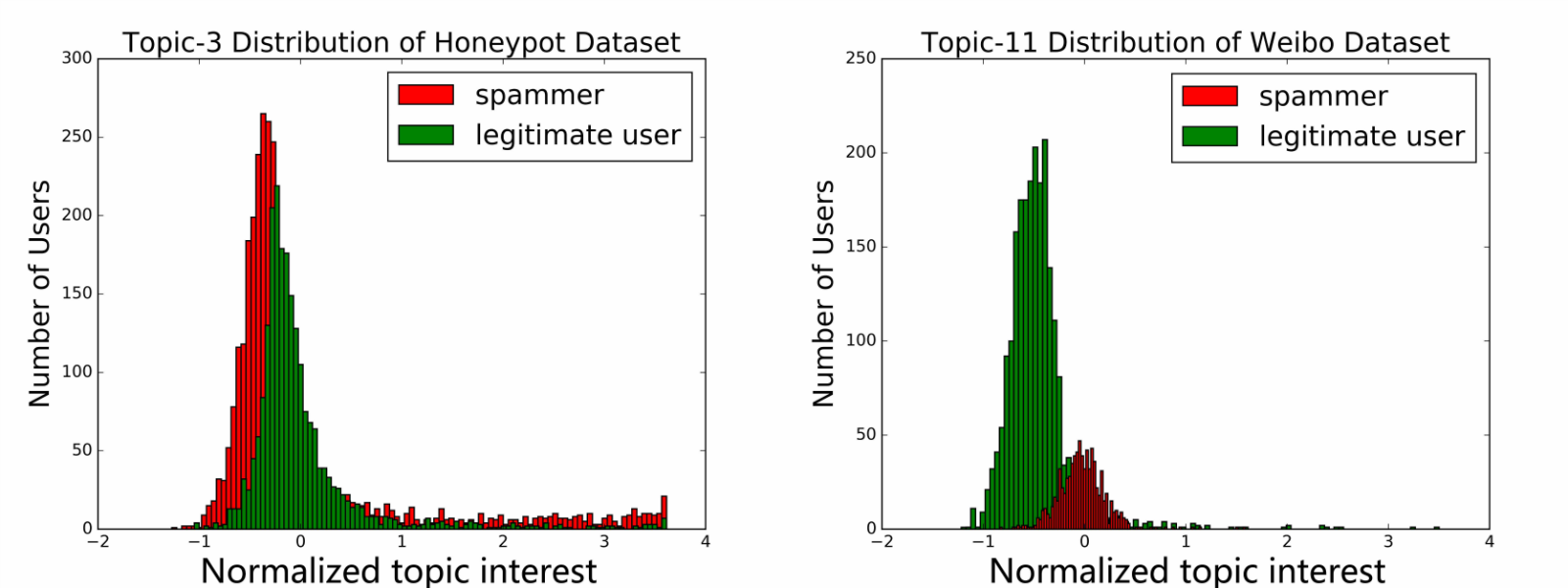}
	\caption[dataset distribution]
	{The topic distribution of legitimate users and social spammers on Honeypot dataset (left) and on Weibo dataset (right), respectively.}\label{visual-dist}

\end{figure}
\begin{itemize}

	\setlength\itemsep{-0.3em}
	\item Legitimate users mainly focus on limited topics which interest him. They seldom post contents unrelated to their concern. 
	\item Content polluters concentrate on certain topics.
	\item Fake accounts focus on a wide range of topics due to random copying and retweeting of other users' tweets.
	\item Spammers and legitimate users show different interests on some topics e.g. commercial, weather, etc.
\end{itemize} 

To better illustrate our observation, Figure.~\ref{visual-dist} shows the topic distribution of spammers and legitimate users in two employed datasets(the Honeypot dataset and Weibo dataset). We can see that on both topics (topic-3 and topic-11) there exists obvious difference between the red bars and green bars, representing spammers and legitimate users. On the Honeypot dataset, spammers have a narrower shape of distribution (the outliers on the red bar tail are not counted) than that of legitimate users. This is because there are more content polluters than fake accounts. In other word, spammers in this dataset tend to concentrate on limited topics.  While on the Weibo dataset, fake accounts who are interested in different topics take large proportion of spammers. Their distribution is more flat (i.e. red bars) than that of the legitimate users. Therefore we can detect spammers by means of the difference of their topic distribution patterns.

\subsection{LDA model}
Blei et al.\shortcite{blei2003latent} first presented Latent Dirichlet Allocation(LDA) as an example of topic model.
\begin{figure}[H]
\includegraphics[clip,scale=0.35]{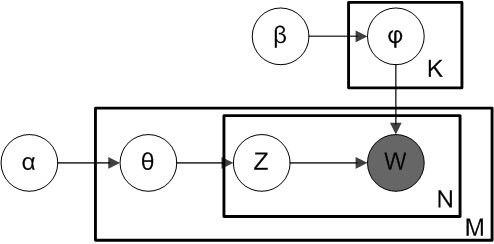}
\caption{The generative model of LDA}\label{lda}
\end{figure}
Each document $i$ is deemed as a bag of words $W=\left\{ w_{i1},w_{i2},...,w_{iM}\right\}$ and $M$ is the number of words. Each word is attributable to one of the document's topics $Z=\left\{ z_{i1},z_{i2},...,z_{iK}\right\}$ and $K$ is the number of topics.
$\psi_{k}$ is a multinomial distribution over words for topic $k$. $\theta_i$ is another multinomial distribution over topics for
document $i$. The smoothed generative model is illustrated in Figure.~\ref{lda}. $\alpha$ and $\beta$ are hyper parameter that affect scarcity of the document-topic and topic-word distributions. 
In this paper, $\alpha$, $\beta$   and $K$ are empirically set to 0.3, 0.01 and 15. 
The entire content of each Twitter user is regarded as one document. We adopt Gibbs Sampling \cite{griffiths2004finding} to speed up the inference of LDA. Based on LDA, we can get the topic probabilities for all users in the employed dataset as:
$X=[X_i;X_2;\cdots;X_n]\in R^{n\times K}$, where $n$ is the number of users. Each element $X_{i}=[p\left(z_{1}\right)p\left(z_{2}\right)\cdots p\left(z_{K}\right)]\in R^{1\times K}$ is a topic probability vector for the $\emph{i}^{th}$ document. $X_i$ is the raw topic probability vector and our features are developed on top of it.

\subsection{Topic-based Features}

Using the LDA model, each person in the dataset is with a topic probability vector $X_i$. Assume $x_{ik}\in X_{i}$ denotes the likelihood that the $\emph{i}^{th}$ tweet account favors $\emph{k}^{th}$ topic in the dataset. Our topic based features can be calculated as below. 

\emph{\textbf{Global Outlier Standard Score}} measures the degree that a user's tweet content is related to a certain topic compared to the other users. Specifically, the "GOSS" score of user $i$ on topic $k$ can be calculated as Eq.(\ref{goss-eq}):

\begin{equation}\label{goss-eq}
\centering
\begin{array}{ll}
\mu\left(x_{k}\right)=\frac{\sum_{i=1}^{n} x_{ik}}{n},\\
GOSS\left(x_{ik}\right)=\frac{x_{ik}-\mu\left(x_k\right)}{\sqrt{\underset{i}{\sum}\left(x_{ik}-\mu\left(x_{k}\right)\right)^{2}}}.
\end{array}
\end{equation}

The value of $GOSS\left(x_{ik}\right)$ indicates the interesting degree of this person to the $\emph{k}^{th}$ topic. Specifically, if $GOSS\left(x_{ik}\right)$ > $GOSS\left(x_{jk}\right)$, it means that the $\emph{i}^{th}$ person has more interest in topic $k$ than the $\emph{j}^{th}$ person. If the value $GOSS\left(x_{ik}\right)$ is extremely high or low, the $\emph{i}^{th}$ person showing extreme interest or no interest on topic $k$ which will probably be a distinctive pattern in the fowllowing classfication. Therefore, the topics interested or disliked by the $\emph{i}^{th}$ person can be manifested by $f_{GOSS}^{i}=[GOSS(x_{i1})\cdots GOSS(x_{iK})]$, from which the pattern of the interested topics with regarding to this person is found. 
Denote $f_{GOSS}^{i}=[GOSS(x_{i1})\cdots GOSS(x_{iK})]$ our first topic-based feature, and it hopefully can get good performance on spammer detection.

\emph{\textbf{Local Outlier Standard Score}} measures the degree of interest someone shows to a certain topic by considering his own homepage content only. For instance, the "LOSS" score of account $i$ on topic $k$ can be calculated as Eq.(~\ref{loss}):

\begin{equation}\label{loss}
\centering
\begin{array}{ll}
\mu\left(x_{i}\right)=\frac{\sum_{k=1}^{K} x_{ik}}{K},\\
LOSS\left(x_{ik}\right)=\frac{x_{ik}-\mu\left(x_i\right)}{\sqrt{\underset{k}{\sum}\left(x_{ik}-\mu\left(x_{i}\right)\right)^{2}}}.
\end{array}
\end{equation}

$\mu(x_i)$ represents the averaged interesting degree for all topics with regarding to $\emph{i}^{th}$ user and his tweet content. Similarly to $GOSS$, the topics interested or disliked by the $\emph{i}^{th}$ person via considering his single post information can be manifested by $f_{LOSS}^{i}=[LOSS(x_{i1})\cdots LOSS(x_{iK})]$, and $LOSS$ becomes our second topic-based features for the $\emph{i}^{th}$ person.
\section{Dataset}

\begin{table*}
	\begin{tabular}{|c|c|c|c|c|c|c|c|}
		\hline 
		\multirow{2}{*}{Feature} & \multirow{2}{*}{Method} & \multicolumn{3}{c|}{Weibo Dataset} & \multicolumn{3}{c|}{Honeypot Dataset}\tabularnewline
		\cline{3-8} 
		&  & Precision & Recall & F1-score & Precision & Recall & F1-score\tabularnewline
		\hline 
		\hline 
		\multirow{3}{*}{GOSS} & SVM & 0.974 & 0.956 & 0.965 & 0.884 & 0.986 & 0.932\tabularnewline
		\cline{2-8} 
		& Adaboost & 0.936 & 0.929 & 0.932 & 0.874 & \textbf{0.990} & 0.928\tabularnewline
		\cline{2-8} 
		& RandomForest & 0.982 & 0.956 & 0.969 & 0.880 & 0.969 & 0.922\tabularnewline
		\hline 
		\multirow{3}{*}{LOSS} & SVM & 0.982 & 0.958 & 0.97 & 0.887 & 0.983 & 0.932\tabularnewline
		\cline{2-8} 
		& Adaboost & 0.941 & 0.929 & 0.935 & 0.878 & 0.976 & 0.924\tabularnewline
		\cline{2-8} 
		& RandomForest & 0.986 & 0.956 & 0.971 & 0.882 & 0.965 & 0.922\tabularnewline
		\hline 
		\multirow{3}{*}{GOSS+LOSS} & SVM & 0.986 & 0.958 & 0.972 & 0.890 & 0.988 & \textbf{0.934}\tabularnewline
		\cline{2-8} 
		& Adaboost & 0.938 & 0.931 & 0.934 & 0.881 & 0.976 & 0.926\tabularnewline
		\cline{2-8} 
		& RandomForest & \textbf{0.988} & \textbf{0.958} & \textbf{0.978} & \textbf{0.895} & 0.951 & 0.922\tabularnewline
		\hline 
	\end{tabular}
	\caption{Performance comparisons for our features with three baseline classifiers}\label{multi-ml-algo}
\end{table*}

We use one public dataset Social Honeypot dataset and one self-collected dataset Weibo dataset to validate the effectiveness of our proposed features.

\textbf{Social Honeypot Dataset}: Lee et al.~\shortcite{lee2010devils} created and deployed 60 seed social accounts on Twitter to attract spammers by reporting back what accounts interact with them. They collected 19,276 legitimate users and 22,223 spammers in their datasets along with their tweet content in 7 months. This is our first test dataset.

\textbf{Our Weibo Dataset}: Sina Weibo is one of the most famous social platforms in China. It has implemented many features from Twitter. The 2197 legitimate user accounts in this dataset are provided by the \emph{Tianchi Competition}\footnote{Tianchi Site http://tianchi.aliyun.com} held by Sina Weibo. The spammers are all purchased commercially from multiple vendors on the Internet. We checked them manually and collected 802 suitable "smart" spammers accounts.

\textbf{Preprocessing}: Before directly performing the experiments on the employed datasets, we first delete some accounts with few posts in the two employed since the number of tweets is highly indicative of spammers. For the English Honeypot dataset, we remove stopwords, punctuations, non-ASCII words and apply stemming. For the Chinese Weibo dataset, we perform segmentation with  "Jieba"\footnote{Jieba Project Page https://github.com/fxsjy/jieba}, a Chinese text segmentation tool. After preprocessing steps, the Weibo dataset contains 2197 legitimate users and 802 spammers, and the honeypot dataset contains 2218 legitimate users and 2947 spammers. It is worth mentioning that the Honeypot dataset has been slashed because most of the Twitter accounts only have limited number of posts, which are not enough to show their interest inclination. 
	\begin{table}[h]
		
		\begin{tabular}{cc|cc}
			&  & \multicolumn{2}{c}{Predicted}\tabularnewline
			&  & Polluter & Legitimate\tabularnewline
			\hline 
			\multirow{2}{*}{Actual} & Polluter & TP & FN\tabularnewline
			& Legitimate & FP & TN\tabularnewline
		\end{tabular}
		\caption{Confusion matrix}\label{confusion}
		\vspace{-0.5cm}
	\end{table}

\section{Experiment}
\subsection{Evaluation Metrics}
The evaluating indicators in our model are show in  \tablename{ \ref{confusion} }. We calculate precision, recall and F1-score (i.e. F1 score) as in Eq.~(\ref{eq:Measure}). Precision is the ratio of selected accounts that are spammers. Recall is the ratio of spammers that are detected so. F1-score is the harmonic mean of precision and recall.
\begin{gather}
precision =\frac{TP}{TP+FP},
recall =\frac{TP}{TP+FN}\nonumber\\
F1-score = \frac{2\times precision \times recall}{precision + recall}\label{eq:Measure}
\end{gather}

%

\begin{table*}
	
	\begin{tabular}{|c|c|c|c|c|c|c|}
		\hline 
		\multirow{2}{*}{Features} & \multicolumn{3}{c|}{SVM} & \multicolumn{3}{c|}{Adaboost}\tabularnewline
		\cline{2-7} 
		& ~Precision~ & ~Recall~ & ~F1-score~ & ~Precision~ & ~Recall~ & ~F1-score~\tabularnewline
		\hline 
		UFN & 0.846 & 0.919 & 0.881 & 0.902 & 0.934 & 0.918\tabularnewline
		\hline 
		UC & 0.855 & 0.904 & 0.879 & 0.854 & 0.901 & 0.877\tabularnewline
		\hline 
		UH & 0.906 & 0.8 & 0.85 & 0.869 & 0.901 & 0.885\tabularnewline
		\hline
		UFN+UC+UH & 0.895 & 0.893 & 0.894 & 0.925 & 0.920 & 0.923\tabularnewline
		\hline 
		LOSS+GOSS & 0.890 & 0.988 & 0.934 & 0.881 & \textbf{0.976} & 0.926\tabularnewline
		\hline 
		UFN+UC+UF+LOSS+GOSS & 0.925 & 0.920 & 0.923 & \textbf{0.952} & 0.946 & \textbf{0.949} \tabularnewline
		\hline 
	\end{tabular}
	\caption{Comparisons of our features and Lee et al.'s features}
\end{table*}

\begin{table}
	
	\begin{tabular}{|c|c|}
		\hline 
		Feature & Description\tabularnewline
		\hline 
		\multirow{4}{*}{UFN} & standard deviation of following \tabularnewline
		\cline{2-2} 
		& standard deviation of followers\tabularnewline
		\cline{2-2} 
		& the number of following\tabularnewline
		\cline{2-2} 
		& following and followers ratio\tabularnewline
		\hline 
		\multirow{4}{*}{UC} & |links| per tweet\tabularnewline
		\cline{2-2} 
		& |@username| in tweets / |tweets|\tabularnewline
		\cline{2-2} 
		& |unique @username| in tweets / |tweets|\tabularnewline
		\cline{2-2} 
		& |unique links| per tweet\tabularnewline
		\hline 
		UH & the change rate of number of following\tabularnewline
		\hline 
	\end{tabular}
	\caption{Honeypot Feature Groups}\label{feature group}
\end{table}

 \subsection{Performance Comparisons with Baseline}
 Three baseline classification methods: Support Vector Machines (SVM), Adaboost, and Random Forests are adopted to evaluate our extracted features. We test each classification algorithm with scikit-learn~\cite{scikit-learn} and run a 10-fold cross validation. On each dataset, the employed classifiers are trained with individual feature first, and then with the combination of the two features. From \tablename{ \ref{multi-ml-algo}}, we can see that GOSS+LOSS achieves the best performance on F1-score among all others. Besides, the classification by combination of LOSS and GOSS can increase accuracy by more than 3\% compared with raw topic distribution probability.

\subsection{Comparison with Other Features}
To compare our extracted features with previously used features for spammer detection, we use three most discriminative feature sets according to Lee et al.~\shortcite{lee2011seven}(\tablename{~\ref{feature group}}). Two classifiers (Adaboost and SVM) are selected to conduct feature performance comparisons. Using Adaboost, our LOSS+GOSS features outperform all other features except for UFN which is 2\% higher than ours with regard to precision on the Honeypot dataset. It is caused by the incorrectly classified spammers who are mostly news source after our manual check. They keep posting all kinds of news pieces covering diverse topics, which is similar to the behavior of fake accounts. However, UFN based on friendship networks is more useful for public accounts who possess large number of followers. The best recall value of our LOSS+GOSS features using SVM is up to 6\% higher than the results by other feature groups. Regarding F1-score, our features outperform all other features. To further show the advantages of our proposed features, we compare our combined LOSS+GOSS with the combination of all the features from Lee et al. ~\shortcite{lee2011seven} (UFN+UC+UH). It's obvious that LOSS+GOSS have a great advantage over UFN+UC+UH in terms of recall and F1-score. Moreover, by combining our LOSS+GOSS features and UFN+UC+UH features together, we obtained another 7.1\% and 2.3\% performance gain with regard to precision and F1-score by Adaboost. Though there is a slight decline in terms of recall. By SVM, we get comparative results on recall and F1-score but about 3.5\% improvement on precision.


 

\section{Conclusion}

In this paper, we propose a novel feature extraction method to effectively detect "smart" spammers who post seemingly legitimate tweets and are thus difficult to identify by existing spammer classification methods. Using the LDA model, we obtain the topic probability for each Twitter user. By utilizing the topic probability result, we extract our two topic-based features: GOSS and LOSS which represent the account with global and local information. Experimental results on a public dataset and a self-built Chinese microblog dataset validate the effectiveness of the proposed features.

\section{Future Work}
In future work, the combination method of local and global information can be further improved to maximize their individual strengths.  We will also apply decision theory to enhancing the performance of our proposed features. Moreover, we are also building larger datasets on both Twitter and Weibo to validate our method.  Moreover, larger datasets on both Twitter and Weibo will be built to further validate our method. 

\bibliography{naacl-submit}
\bibliographystyle{naaclhlt2016}

\end{document}